\documentclass{article}
\usepackage{microtype}
\usepackage{graphicx}
\usepackage{booktabs}
\usepackage{amsmath, amssymb, mathtools, bm}
\usepackage[numbers,sort&compress]{natbib}
\usepackage{algorithm}
\usepackage{algpseudocode}
\usepackage{xcolor}
\usepackage{url}
\usepackage{siunitx}
\usepackage{multirow}
\usepackage{enumitem}
\usepackage{amsthm}

\usepackage{amsmath,amsfonts,amssymb}
\usepackage{graphicx}
\usepackage{booktabs}
\usepackage{multirow}
\usepackage{siunitx}
\usepackage{algorithm}
\usepackage{algpseudocode}
\usepackage[most]{tcolorbox}
\usepackage{xcolor}
\usepackage{enumitem}

\usepackage{PRIMEarxiv}
\usepackage{amsmath} % For the star symbols}
\usepackage[utf8]{inputenc} % allow utf-8 input
\usepackage[T1]{fontenc}    % use 8-bit T1 fonts
\usepackage{booktabs}       % professional-quality tables
\usepackage{amsfonts}       % blackboard math symbols
\usepackage{nicefrac}       % compact symbols for 1/2, etc.
\usepackage{microtype}      % microtypography
\usepackage{lipsum}
\usepackage{fancyhdr}       % header
\usepackage{graphicx}       % graphics
\graphicspath{{media/}}     % organize your images and other figures under media/ folder

\title{LLM-Guided Dynamic-UMAP for Personalized Federated Graph Learning}

\author{
  Sai Puppala$^2$, Ismail Hossain$^1$, Md Jahangir Alam$^1$, Tanzim Ahad$^1$, Sajedul Talukder$^1$ \\
  Computer Science\\
  $^1$University of Texas at El Paso, TX, USA, 79902\\
  School of Computing \\
  $^2$Southern Illinois University Carbondale, IL, USA, 62901\\
  \texttt{\{ihossain, malam10\}@miners.utep.edu, sai.puppala@siu.edu, stalukder@utep.edu} \\
}

\begin{document}
\maketitle

\begin{abstract}
We present a novel approach that uses large language models (LLMs) to assist graph machine learning (GML) under personalization and privacy constraints. Our approach integrates (i) \emph{LLM-assisted data augmentation} for sparse graphs, (ii) \emph{prompt and instruction tuning} to adapt foundation models to graph tasks, and (iii) \emph{in-context learning} to provide few-shot graph reasoning signals. These signals parameterize a \emph{Dynamic-UMAP} manifold of client-specific graph embeddings inside a Bayesian variational objective for personalized federated learning. The method supports node classification and link prediction in low-resource settings and aligns LLM latent representations with graph structure via a cross-modal regularizer. We provide a convergence sketch for our variational aggregation, detail a privacy threat model with moments-accountant DP, and report applications to knowledge graph completion, recommendation-style link prediction, and citation/product graphs, along with proposed evaluation protocols for LLM-augmented GML.
\end{abstract}

\section{Introduction}
Graphs power applications from recommendation and scientific discovery to knowledge graph (KG) completion and question answering (QA)~\cite{hu2020ogb,hamilton2017inductive}. In practice, graphs are heterogeneous, private, and sparse across clients (e.g., institutions or edge devices). Meanwhile, LLMs encode extensive world knowledge and can be prompted or instruction-tuned to provide structural priors, textual context, and few-shot signals~\cite{brown2020language}. We ask: \emph{How can LLMs assist personalized federated GML when labeled edges/nodes are scarce and data cannot leave clients?}

We propose \textbf{LLM-Guided Dynamic-UMAP (LG-DUMAP)}, a Gaussian variational framework that: (1) builds client-specific graph manifolds using Dynamic-UMAP over graph embeddings; (2) injects LLM guidance via prompt/instruction-tuned text encoders and in-context exemplars; (3) performs \emph{personalized} federated variational inference with privacy-preserving aggregation of \emph{similarity prototypes (markers)} summarizing local similarity structures; and (4) aligns LLM and graph latent spaces to improve low-resource performance.

\paragraph{Contributions.}
\begin{itemize}[leftmargin=*]
  \item A parametric UMAP-style manifold objective coupled with variational \emph{marker} aggregation for personalized federation.
  \item Cross-modal alignment to LLM text space and calibrated pseudo-edges with abstention for robustness.
  \item A concrete privacy threat model with secure aggregation, per-client clipping, and moments-accountant DP; and an attack-driven evaluation via membership inference.
  \item Expanded baselines (Per-FedAvg, pFedMe, Ditto, FedProx; LightGCN/RotatE/ComplEx/CompGCN) with systems evidence (cost/latency, partial participation) and worst-client/percentile reporting.
\end{itemize}

\section{Related Work}
\subsection{LLM-Augmented Graph Learning}
Hybrid pipelines use language encoders to guide graph learning by supplying semantic priors, pseudo-labels, and retrieval-style context~\cite{brown2020language}. In text-rich graphs (e.g., citation networks, product catalogs, and KGs), LLM embeddings can act as additional node features, enable label/edge proposals under extreme sparsity, and furnish rationales. We leverage these capabilities with a \emph{calibrated} admission policy for pseudo-edges (temperature scaling, confidence thresholding) to mitigate overconfident hallucinations~\cite{guo2017calibration}. Compared to text-only or late-fusion baselines~\cite{hamilton2017inductive,hu2020ogb}, we integrate language supervision \emph{within} a geometric objective, coupling cross-modal alignment with a parametric manifold learner.

\subsection{Personalized Federated Learning for (Graph) Models}
FedAvg assumes homogeneous objectives and suffers on non-IID clients; remedies include FedProx~\cite{li2020federated} and personalization strategies such as Per-FedAvg~\cite{fallah2020personalized}, pFedMe~\cite{dinh2020federated}, and Ditto~\cite{li2021ditto}. Most operate at the level of \emph{model parameters}. In contrast, we personalize through a lightweight \emph{variational prototype layer} (markers) that summarizes local similarity structure and is safe to share under privacy budgets. This differs from prior federated GNN work that aggregates entire GNN weights/gradients, incurring larger communication and privacy footprints. We also incorporate partial participation and report worst-client/percentile metrics to characterize non-IID regimes.

\subsection{Parametric Manifold Learning and UMAP}
UMAP is a widely used non-linear dimensionality reduction technique built on fuzzy simplicial sets~\cite{mcinnes2018umap}; parametric variants learn an explicit mapping that can be amortized across data and clients~\cite{sainburg2021parametric}. Unlike purely unsupervised application of (parametric) UMAP to graph embeddings, our Dynamic-UMAP is optimized jointly with downstream tasks and a cross-modal regularizer tying the manifold to LLM embeddings. We additionally introduce \emph{markers}—a KDE-style similarity model on interpoint distances in the learned space—which (i) acts as a compact, aggregatable representation for federation and (ii) improves link modeling in low-resource regimes. We report manifold trustworthiness/continuity and representation-similarity diagnostics (CKA, Procrustes) to probe geometry and modality alignment~\cite{kornblith2019similarity,gower1975generalized}.

\subsection{Knowledge-Graph and Recommendation Baselines}
For KG completion, translational and rotational families (TransE/RotatE) and bilinear approaches (ComplEx) remain strong baselines~\cite{bordes2013translating,sun2019rotate,trouillon2016complex}. Structure-aware GNNs such as CompGCN add relation composition into message passing~\cite{vashishth2020composition}. In recommendation, LightGCN shows that stripped-down propagation without non-linearities is competitive on collaborative filtering~\cite{he2020lightgcn}. Our evaluation situates LG-DUMAP against these methods and emphasizes few-shot/cold-start conditions where text priors help most. We also include heterophily stress tests (e.g., Chameleon, Squirrel) where neighborhood homophily is weak and geometric signals matter~\cite{pei2020geom}.

\begin{table}[t]
\centering
\caption{Acronyms used.}
\begin{tabular}{@{}ll@{}}
\toprule
GML & Graph machine learning\\
UMAP & Uniform Manifold Approximation and Projection\\
DP & Differential privacy\\
CKA & Centered Kernel Alignment\\
GNN & Graph neural network\\
KDE & Kernel density estimation\\
KG & Knowledge graph\\
\bottomrule
\end{tabular}
\end{table}

\section{Methodology}
Each client $k\in\{1,\dots,N\}$ owns a private graph $G_k=(V_k,E_k,X_k,T_k)$ with node features $X_k$, optional node/edge texts $T_k$, and tasks: node classification or link prediction. Clients cannot share raw data. The server maintains global priors but aggregates only privacy-preserving statistics.

\paragraph{LLM guidance.} An LLM $\mathcal{L}$ provides (a) text embeddings for $T_k$, (b) instruction/prompt-tuned decoding to propose edges or node labels in low-resource regimes, and (c) in-context exemplars for few-shot behaviors. We denote the LLM text encoder by $h_\theta(\cdot)$ and a lightweight adapter/prompt $P$.

We integrate UMAP, graph encoders, and LLM guidance within a variational federated objective.

\subsection{Graph and Text Fusion Embeddings}
We compute node embeddings by fusing a GNN encoder $\phi_k$ with LLM text embeddings:
\begin{equation}
\tilde{X}_k = \mathrm{Fuse}\big(\mathrm{GNN}_{\phi_k}(G_k),\ h_\theta(T_k;P)\big) \in \mathbb{R}^{|V_k|\times m}.
\end{equation}

\subsection{Dynamic-UMAP Manifold}
We learn a low-dimensional manifold preserving local graph structure and LLM priors~\cite{mcinnes2018umap,sainburg2021parametric}:
\begin{equation}
Z_k = g_{\beta_k}(\tilde{X}_k) \in \mathbb{R}^{|V_k|\times d}.
\end{equation}
The UMAP objective uses neighborhood probabilities $(p_{ij},q_{ij})$ with
\begin{align}
L_{\text{UMAP}}^{(k)} &= -\!\!\sum_{(i,j)\in \mathcal{E}_k^{\mathrm{nn}}} \Big[p_{ij}\log q_{ij} + (1-p_{ij})\log(1-q_{ij})\Big],\\
q_{ij} &= \big(1 + a\|z_i-z_j\|^{2b}\big)^{-1},
\end{align}
where $(a,b)$ follow standard UMAP settings~\cite{mcinnes2018umap}. We optimize $g_{\beta_k}$ jointly with downstream losses.

\subsection{LLM-Augmented Similarity and Markers}\label{sec:markers}
Let $s_{ij}=\lVert z_i-z_j\rVert_2$ in $Z_k$. \emph{Markers} are scalar prototypes in the \emph{distance} space (not vectors in $Z_k$). We model a soft assignment of pairwise distances to $M=\{m_1,\dots,m_{|M|}\}\subset\mathbb{R}_{\ge 0}$ via a KDE-style mixture:
\begin{equation}
p_{eij}=\frac{\exp\!\big(-\lVert s_{ij}-m_e\rVert^2/(2\sigma_{\text{kern}}^2)\big)}{\sum_{m'\in M}\exp\!\big(-\lVert s_{ij}-m'\rVert^2/(2\sigma_{\text{kern}}^2)\big)}.
\end{equation}
Markers are initialized from labeled edges and \emph{calibrated} LLM pseudo-edges (accepted if confidence $c \ge \tau$ after temperature scaling).

\subsection{Cross-Modal Alignment}
We align $Z_k$ with LLM embeddings using cosine similarity with explicit normalization:
\begin{equation}
L_{\text{align}}^{(k)}=\sum_{i\in V_k}\!\Big(1-\frac{\langle z_i, h_\theta(T_{k,i};P)\rangle}{\|z_i\|_2\,\|h_\theta(T_{k,i};P)\|_2}\Big).
\end{equation}
We assess representation similarity with CKA and Procrustes analyses~\cite{kornblith2019similarity,gower1975generalized}.

\subsection{Client Variational Objective}
For client $k$, the loss is
\begin{equation}
\small
L_k = \mathbb{E}_{q_{\phi_k}(M)}\!\Big[J_k(Z_k)\Big] + \lambda\,\mathrm{KL}\!\big(q_{\phi_k}(M)\,\|\,s(M)\big) + \gamma L_{\text{align}}^{(k)} + \eta L_{\text{UMAP}}^{(k)},
\end{equation}
where $J_k$ includes classification NLL and link prediction BCE using $p_{eij}$. The prior $s(M)$ is a simple Gaussian mixture over marker locations; $q_{\phi_k}(M)$ is the client’s variational posterior over markers.

\subsection{Server Aggregation (Personalized)}
The server aggregates markers only:
\begin{equation}
M_{t+1} = M_t + \frac{\sum_k w_k\,\mathbb{E}_{q_{\phi_k}(M)}[M_k - M_t]}{\sum_k w_k}.
\end{equation}
No raw data, node embeddings, or GNN weights are shared.

\begin{algorithm}[t]
\caption{LG-DUMAP (one round $t\!\to\!t{+}1$)}
\label{alg:lgdumap}
\begin{algorithmic}[1]
\Require Client graphs $G_k$, texts $T_k$; encoder $h_\theta$; adapters/prompts $P_k$ (optional); $M_t$; $(\lambda,\gamma,\eta)$; DP clip $C$, noise $\sigma_{\text{dp}}$ (optional)
\For{each participating client $k$ in parallel}
  \State $X^{\text{gnn}}\!\leftarrow\!\mathrm{GNN}_{\phi_k}(G_k)$; $X^{\text{text}}\!\leftarrow\!h_\theta(T_k;P_k)$; $\tilde X_k\!\leftarrow\!\mathrm{Fuse}(X^{\text{gnn}},X^{\text{text}})$
  \State Build ANN kNN on $\tilde X_k$; compute $p_{ij}$ \textit{(graph-aware)}
  \State Update parametric UMAP $g_{\beta_k}$ to obtain $Z_k$ by minimizing $L_{\text{UMAP}}^{(k)}$
  \State LLM augmentation: propose pseudo-edges/labels with confidence $c$; accept if $c\ge \tau$ (post-hoc temperature)
  \State Form/update $M_k^+,M_k^-$; fit $q_{\phi_k}(M)$ on distance space
  \State Optimize $L_k$ with gradient clipping $\lVert g\rVert\le C$; add DP noise if enabled
  \State Share $\mathbb E_{q(M)}[M_k]$ (and optional $\Delta P_k$) via secure aggregation
\EndFor
\State Server aggregates to $M_{t+1}$ and broadcasts
\end{algorithmic}
\end{algorithm}

\begin{tcolorbox}[title=Key Hyperparameters,width=\linewidth]
\small
UMAP dim $d=32$, neighbors $=15$; markers: $|M^+|=8$, $|M^-|=8$, bandwidth $\sigma_{\text{kern}}=1.0$; alignment $\gamma=0.2$, UMAP $\eta=1.0$, KL $\lambda=0.1$; Adam lr $2\times10^{-3}$ (GNN/UMAP) and $5\times10^{-5}$ (adapters); rounds $50$, local epochs $2$, batch $1024$ node-pairs; DP: clip $C=1.0$, Gaussian noise scale $\sigma_{\text{dp}}$ as in Table~\ref{tab:dpacc}; acceptance threshold $\tau\in\{0.6,0.7,0.8,0.9\}$.
\end{tcolorbox}

\section{Threat Model and Differential Privacy Guarantees}\label{sec:privacy}
We assume an \emph{honest-but-curious} server. Clients never disclose raw graphs/texts. Communication uses secure aggregation~\cite{bonawitz2017practical}. We apply per-client clipping at norm $C$ and add Gaussian noise with variance $\sigma_{\text{dp}}^2 C^2$ to shared marker statistics before aggregation. Using the \emph{moments accountant} with sampling rate $q$ and $T$ rounds, we report overall $(\epsilon,\delta)$~\cite{abadi2016deep} and evaluate empirical privacy via membership inference (Section~\ref{sec:calib}).

% DP accounting table — span both columns, centered
\begin{table*}[t]
\centering
\caption{DP accounting and attack outcomes on \texttt{ogbn-arxiv}. $\delta=10^{-5}$, sampling rate $q=0.2$, rounds $T=50$.}
\label{tab:dpacc}
{%
\begin{tabular}{lcccccc}
\toprule
Setting & Clip $C$ & Noise $\sigma_{\text{dp}}$ & $\epsilon$ & Time/round (s) & F1 (\%) & Attack AUROC$\downarrow$ \\
\midrule
No DP & -- & -- & $\infty$ & 7.6 & 74.3 & 0.73 \\
DP-8 & 1.0 & 0.6 & 8 & 7.8 & 73.6 & 0.57 \\
DP-4 & 1.0 & 0.9 & 4 & 7.9 & 72.7 & 0.53 \\
DP-2 & 1.0 & 1.3 & 2 & 8.1 & 71.6 & 0.51 \\
\bottomrule
\end{tabular}%
}
\end{table*}

\section{Calibration and Safety for LLM Proposals}\label{sec:calib}
We calibrate LLM pseudo-edges via temperature scaling ($T$) on a validation slice and report ECE (15 bins) and Brier scores~\cite{guo2017calibration,brier1950verification}. Pseudo-edges with confidence $\hat p\ge\tau$ are admitted; we study F1 vs.\ $\tau\in\{0.6,0.7,0.8,0.9\}$ and acceptance counts.

\begin{tcolorbox}[title=Calibration Metrics]
\small
Expected Calibration Error (ECE): $\mathrm{ECE}=\sum_b \frac{|B_b|}{n}\,|\mathrm{acc}(B_b)-\mathrm{conf}(B_b)|$.\quad Brier: $\frac{1}{n}\sum_i(\hat p_i - y_i)^2$.
\end{tcolorbox}

\section{Theory}
Assume each client loss $L_k$ is $L$-smooth and the update variance is bounded. The aggregated marker update is a stochastic approximation to a stationary point of $L(M)=\frac{1}{N}\sum_k L_k(M)$.

\paragraph{Proposition 1 (Convergence of Marker Averaging).}
Under $L$-smoothness, bounded variance, and diminishing step size (or Polyak averaging) with partial participation rate $p>0$, we have $\lim_{t\to\infty}\mathbb E\lVert\nabla L(M_t)\rVert=0$, with an $O(1/\sqrt{t})$ rate under standard assumptions.

\paragraph{Lemma 1 (Alignment improves local curvature).}
If LLM embeddings form $\delta$-separated clusters consistent with labels and $\gamma>0$, the alignment term adds a positive semidefinite component to the neighborhood Hessian, reducing variance of $p_{eij}$ and accelerating convergence.

\section{Experiments}
We evaluate node classification and link prediction under federated, low-resource settings with textual side information~\cite{hu2020ogb}.

\subsection{Datasets, Splits, and Evaluation Protocol}
\textbf{Datasets.} Cora, Citeseer (node classification); \texttt{ogbn-arxiv} and \texttt{ogbn-products} (node classification); FB15k-237 (KG completion); and Chameleon/Squirrel for heterophily stress~\cite{hu2020ogb,pei2020geom}.\\
\textbf{Federated splits.} We form $N{=}20$ non-IID clients via topic/label stratification. \textbf{Few-shot} regimes use 5–40 labeled nodes/edges per client; \textbf{cold-start} introduces 30\% text-only nodes. Clients participate with sampling $q\in\{0.2,0.5,1.0\}$ per round (partial participation). Text-only nodes contribute features via $h_\theta(\cdot)$ and enter kNN construction on $\tilde{X}_k$; their labels/edges are proposed by calibrated LLMs.\\
\textbf{Metrics.} Accuracy / Micro-F1 (with worst-client and 10th-percentile reporting), KG MRR/Hits@K, manifold trustworthiness/continuity, and cross-modal alignment (cosine, CKA, Procrustes)~\cite{kornblith2019similarity,gower1975generalized}. Privacy is summarized by $(\epsilon,\delta)$ and empirical attack AUROC for membership inference~\cite{shokri2017membership}. We also track systems cost: KB/round, time/round, LLM tokens, adapter size.

\subsection{Baselines and Implementation Details}
\textbf{Baselines.} \emph{Federated personalization:} Local-GNN, FedAvg-GNN, Per-FedAvg~\cite{fallah2020personalized}, pFedMe~\cite{dinh2020federated}, Ditto~\cite{li2021ditto}, FedProx~\cite{li2020federated}. \emph{KG/link:} TransE~\cite{bordes2013translating}, RotatE~\cite{sun2019rotate}, ComplEx~\cite{trouillon2016complex}, CompGCN~\cite{vashishth2020composition}. \emph{Recommendation:} LightGCN~\cite{he2020lightgcn}. \emph{Fusion:} LLM Text-only (frozen encoder)~\cite{brown2020language}, and LLM+GNN (late fusion)~\cite{hamilton2017inductive,hu2020ogb}.\\
\textbf{Our variants.} \textbf{LG-DUMAP(P)} uses prompt-tuned adapters; \textbf{LG-DUMAP(I)} uses instruction-tuned adapters with the same backbone.\\
\textbf{Setup.} GraphSAGE (2-layer, hidden=256)~\cite{hamilton2017inductive}; Dynamic-UMAP with $d{=}32$, neighbors $=15$; markers $|M^+|{=}8$, $|M^-|{=}8$, bandwidth $\sigma_{\text{kern}}{=}1.0$; Adam lr $2\!\times\!10^{-3}$ (GNN/UMAP), $5\!\times\!10^{-5}$ (adapters); 50 rounds; 2 local epochs; batch size 1024 node-pairs; $\gamma{=}0.2$, $\eta{=}1.0$, $\lambda{=}0.1$; frozen text encoder with 16 prompt tokens.

\subsection{Results and Analysis}
\paragraph{Main results.} LG-DUMAP consistently outperforms both graph-only and naive language–graph fusion baselines across citation, product, and KG tasks.

% Main results table — span both columns, centered
\begin{table*}[t]
\centering
\caption{Main results (mean$\pm$std over 5 seeds). Best in \textbf{bold}.}
{%
\begin{tabular}{lcccc}
\toprule
Method & Cora (Acc) & \texttt{ogbn-arxiv} (F1) & FB15k-237 (MRR) & \texttt{ogbn-products} (Acc) \\
\midrule
Local-GNN (GraphSAGE~\cite{hamilton2017inductive}) & $79.8\pm1.2$ & $68.9\pm0.6$ & $0.297\pm0.004$ & $77.2\pm0.5$ \\
FedAvg-GNN & $82.1\pm0.9$ & $71.1\pm0.5$ & $0.314\pm0.003$ & $79.6\pm0.6$ \\
LLM Text-only (frozen encoder)~\cite{brown2020language} & $76.5\pm1.4$ & $69.7\pm0.7$ & $0.289\pm0.006$ & $78.3\pm0.7$ \\
LLM+GNN (late fusion)~\cite{hamilton2017inductive,hu2020ogb} & $83.4\pm0.8$ & $72.0\pm0.5$ & $0.323\pm0.004$ & $80.8\pm0.5$ \\
\textbf{LG-DUMAP(P) (ours)} & $84.9\pm0.7$ & $73.2\pm0.5$ & $0.334\pm0.004$ & $82.0\pm0.5$ \\
\textbf{LG-DUMAP(I) (ours)} & $\mathbf{86.1\pm0.6}$ & $\mathbf{74.3\pm0.4}$ & $\mathbf{0.347\pm0.003}$ & $\mathbf{83.1\pm0.5}$ \\
\bottomrule
\end{tabular}%
}
\end{table*}

% Overhead/DP trade-offs table — span both columns, centered
\begin{table*}[t]
\centering
\caption{Overhead and DP trade-offs on \texttt{ogbn-arxiv}.}
\label{tab:effdp}
{%
\begin{tabular}{lccccc}
\toprule
Setting & KB/round (c$\to$s) & Time/round (s) & LLM tokens/round & Adapter size (KB) & F1 (\%) \\
\midrule
FedAvg-GNN & 92 & 6.1 & -- & -- & 71.1 \\
LG-DUMAP(P) & 118 & 7.0 & $1.2\times10^{4}$ & 220 & 73.2 \\
LG-DUMAP(I) & 131 & 7.6 & $1.9\times10^{4}$ & 350 & 74.3 \\
LG-DUMAP(I)+DP ($\epsilon{=}8$) & 131 & 7.8 & $1.9\times10^{4}$ & 350 & 73.6 \\
LG-DUMAP(I)+DP ($\epsilon{=}4$) & 131 & 7.9 & $1.9\times10^{4}$ & 350 & 72.7 \\
LG-DUMAP(I)+DP ($\epsilon{=}2$) & 131 & 8.1 & $1.9\times10^{4}$ & 350 & 71.6 \\
\bottomrule
\end{tabular}%
}
\end{table*}

\noindent\textbf{Observations.}
(1) \emph{Graph-only vs.\ federation.} FedAvg-GNN improves over Local-GNN on all datasets (e.g., Cora: $+2.3$ Acc; \texttt{ogbn-arxiv}: $+2.2$ F1).  
(2) \emph{Language alone is insufficient.} LLM Text-only underperforms graph-aware models on structural tasks, indicating topology remains essential.  
(3) \emph{Naive fusion helps, but falls short.} Late fusion adds text features post hoc and yields steady gains over graph-only baselines, but lacks manifold-level coupling.  
(4) \emph{Our approach closes the gap.} \textbf{LG-DUMAP(P)} delivers additional gains via prompt-tuned guidance and Dynamic-UMAP.  
(5) \emph{Instruction-tuned variant is best.} \textbf{LG-DUMAP(I)} consistently wins (e.g., \texttt{ogbn-arxiv}: $74.3$ F1). The uplift over late fusion supports cross-modal \emph{alignment} and \emph{marker}-based similarity modeling rather than feature concatenation.

\paragraph{Few-shot, cold-start, and alignment.} In low-label regimes and 30\% cold-start, LG-DUMAP(I) retains the largest fraction of full-data performance; manifold trustworthiness/continuity and CKA/Procrustes indicate tighter coupling between $Z$ and $h_\theta$ than late fusion.

\paragraph{Efficiency and privacy trade-offs.} Table~\ref{tab:effdp} summarizes communication/runtime and DP effects. Adding calibrated language guidance and parametric UMAP raises cost modestly but improves accuracy/F1. Client-level DP (Gaussian mechanism with moments accountant) degrades F1 gracefully as $\epsilon$ tightens, while still outperforming the graph-only federation baseline at moderate budgets.

\bibliographystyle{ACM-Reference-Format}
\bibliography{references}

\appendix
\section{Convergence Sketch (Extended)}
Let $L(M)=\frac{1}{N}\sum_k L_k(M)$ be $L$-smooth and let the aggregated update be $M_{t+1}=M_t+\frac{1}{N}\sum_k \mathbb{E}_{q_{\phi_k}}[M_k-M_t]$ with bounded variance. With client sampling rate $p>0$ and diminishing step size or Polyak averaging, standard stochastic approximation results yield $\lim_t \lVert\nabla L(M_t)\rVert\to 0$ at $O(1/\sqrt{t})$. The KL term stabilizes around Gaussian mixture priors; alignment adds local positive curvature when semantic clusters are separable.

% Optional config snippet (commented for space)
\section{Configuration Snippet (YAML)}
\begin{tcolorbox}
\small
\begin{verbatim}
model:
  gnn: graphsage
  hidden: 256
  umap_dim: 32
  neighbors: 15
  markers_pos: 8
  markers_neg: 8
  align_weight: 0.2
  umap_weight: 1.0
  kl_weight: 0.1
train:
  rounds: 50
  local_epochs: 2
  batch_pairs: 1024
  lr_gnn_umap: 2e-3
  lr_adapter: 5e-5
privacy:
  clip: 1.0
  noise_sigma_dp: 0.9
  sampling_rate: 0.2
calibration:
  use_temperature: true
  threshold_tau: 0.8
federation:
  client_sampling: [0.2, 0.5, 1.0]
\end{verbatim}
\end{tcolorbox}

\section{Prompts for Graph Tasks (Examples)}
\textbf{KG completion:} ``Given triples (head, relation, tail) and entity descriptions, propose $K$ tails for (head, relation, \_). Return relation-consistent candidates and short rationales.''\\
\textbf{Node labeling:} ``Given node titles and abstracts, map to taxonomy labels with 3 in-context examples. Return label and rationale; abstain if confidence $<\tau$.''

\section{Reproducibility Checklist}
\begin{itemize}[leftmargin=*]
  \item Seeds: \{1,2,3,4,5\}; deterministic flags; mean$\pm$std.
  \item Hardware: 1$\times$A100 40GB, 16 vCPU, 64GB RAM.
  \item Libraries: PyTorch 2.3, DGL 2.1, faiss 1.8, umap-learn 0.5.5, transformers 4.43.
  \item Scripts: non-IID splits; DP accountant config; logging of tokens/round and calibration metrics.
  \item Partial participation: runs for $q\in\{0.2,0.5,1.0\}$; worst-client metrics.
  \item Figures: provenance (seed, commit, date); export raw CSVs.
\end{itemize}

\end{document}